\documentclass{article}
\usepackage{spconf,amsmath,graphicx,hyperref}

\title{Unsupervised Integrated-Circuit Defect Segmentation via Image-Intrinsic Normality}
\name{Botong.Zhao, Qijun.Shi, Shujing.Lyu, Yue.Lu*\thanks{Thanks to XYZ agency for funding.}}
\address{Shanghai Key Laboratory of Multidimensional Information Processing, East China Normal University}
\begin{document}
%
\maketitle
\begin{abstract}
Modern Integrated-Circuit(IC) manufacturing introduces diverse, fine-grained defects that depress yield and reliability. Most industrial defect segmentation compares a test image against an external normal set, a strategy that is brittle for IC imagery where layouts vary across products and accurate alignment is difficult. We observe that defects are predominantly local, while each image still contains rich, repeatable normal patterns. We therefore propose an unsupervised IC defect segmentation framework that requires no external normal support. A learnable normal-information extractor aggregates representative normal features from the test image, and a coherence loss enforces their association with normal regions. Guided by these features, a decoder reconstructs only normal content; the reconstruction residual then segments defects. Pseudo-anomaly augmentation further stabilizes training. Experiments on datasets from three IC process stages show consistent improvements over existing approaches and strong robustness to product variability.
\end{abstract}
\section{Introduction}
\label{sec:intro}

The fabrication of IC involves hundreds of steps, such as dummy poly removal, etching, and chemical-mechanical polishing. Each step may introduce various defects. Efficient management and control of wafer defects is critical for yield improvement. However, scanning electron microscope (SEM) images of wafer surfaces exhibit complex backgrounds, an extremely low prevalence of true defects, and highly diverse shapes.

In recent years, industrial defect segmentation has achieved substantial progress. Memory-bank approaches leverage pretrained representations of normal samples for contrastive matching to segment anomalies \cite{li2024musc, jeong2023winclip, zuo2024reconstruction}. Reconstruction-based methods model the distribution of normal appearance and use reconstruction residuals as defect indicators \cite{fuvcka2024transfusion, yao2024glad, he2024diffusion}. With the emergence of multimodal models, prompt-driven vision–language systems have also been explored for anomaly detection \cite{ma2025aa, gu2024anomalygpt, li2024promptad}. However, as illustrated in Figure 1, unlike conventional industrial products that typically exhibit uniform appearance, SEM images of IC are characterized by highly structured and repetitive patterns. Defects in such images often manifest as subtle, fine-grained local anomalies embedded within the background, posing a unique challenge for generic visual models. This intrinsic discrepancy with generic pretrained models, which emphasize global semantic cues, poses significant challenges to the stability and transferability of existing methods. 
Lu employs Masked Autoencoder-based reconstruction to segment defects in IC SEM images \cite{lu2023masked}; Chen proposes hybrid CNN-Transformer frameworks for IC defect segmentation \cite{mei2025novel,qiao2024deepsem}; and Jiang explores trainable multimodal large models to align IC defect semantics \cite{jiang2024fabgpt}. Nevertheless, these approaches still rely on support-set normal exemplars and require substantial manual annotation, and their generalization across diverse layouts and process stages remains difficult to guarantee.

\begin{figure}[ht]
        \centering
	\includegraphics[width=0.45\textwidth]{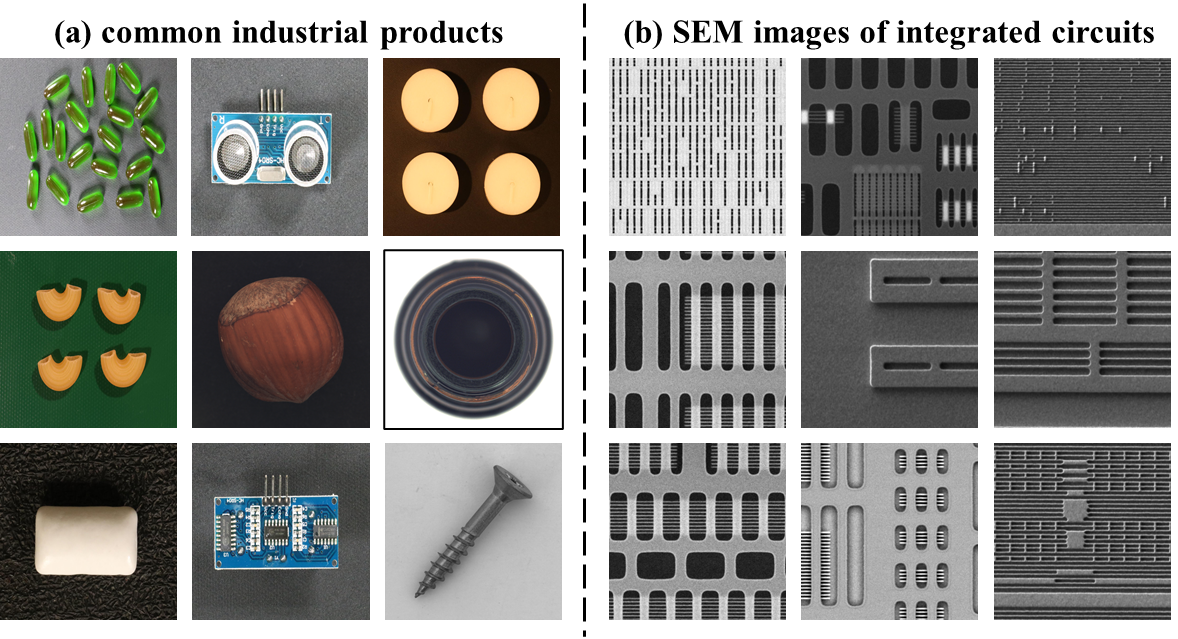}
	\caption{Differences between industrial defects and IC defects.}
	\label{Fig:1}       
\end{figure}

\begin{figure*}[ht]
        \centering
	\includegraphics[width=0.9\textwidth]{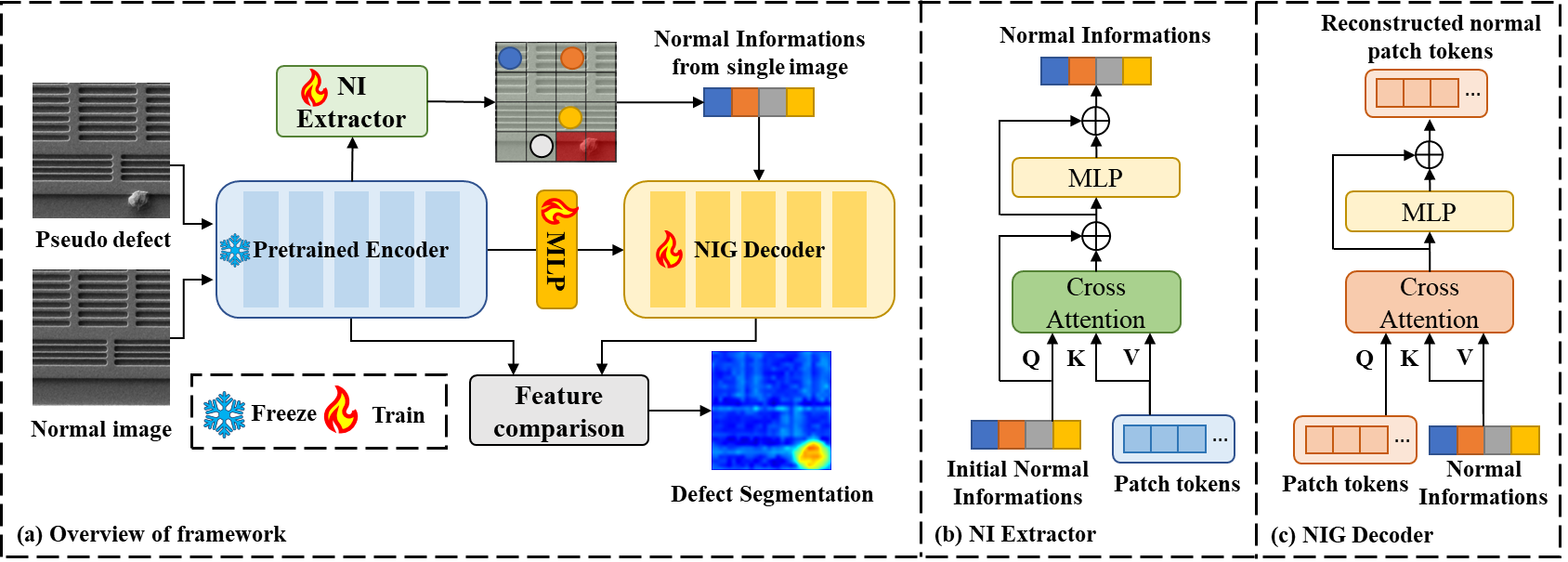}
	\caption{(a) Overview of IC defect segmentation via Image-Intrinsic Normality. (b)Detailed architecture of each layer in the NI Extractor. (c)Detailed architecture of each layer in the NIG Decoder.}
	\label{Fig:2}       
\end{figure*}

Motivated by the above observation, we note that even defective IC images still contain structured and repetitive normal regions. This insight inspires us to eliminate the reliance on external normal references and instead mine normal information directly from the test image itself to guide defect segmentation. We propose an unsupervised IC defect segmentation framework. Specifically, we design a learnable Normal-Information(NI) Extractor that is trained using both normal samples and synthetically generated abnormal samples, enabling it to extract representative Normal Informations from the image. A consistency loss is introduced to ensure that the extracted features are associated with truly normal regions. Importantly, since these features describe the characteristics of normal areas rather than exact pixel-wise patterns, we avoid direct feature comparison. Instead, we design a Normal-Information-Guided (NIG) Decoder to reconstruct the features of local regions based on the extracted Normal Informations. The discrepancy between the reconstructed and original regions is then used to segment defects. This strategy enhances both the robustness and sensitivity of our method.

Our main contributions are summarized as follows:
1.	We propose an unsupervised IC defect segmentation method that leverages the normal information inherently present within the test image itself, eliminating the need for external reference samples; 
2.	We introduce a reconstruction strategy guided by extracted normal features. Instead of directly comparing feature distances, our method reconstructs local regions and identifies defects based on residual differences, resulting in more stable and robust localization; 
3.	We evaluate our approach on datasets from three different IC process stages and demonstrate superior performance over existing methods, validating the effectiveness and generalizability of our framework.

\section{Method}
\label{sec:Method}
Existing methods rely on normal features from a support set and compare them against the test image. However, due to the complex backgrounds of IC, these normal features are difficult to directly align with the local features of the test image, especially under the interference of global semantics and positional encodings. As illustrated in Figure 2(a), our framework consists of two main stages. First, we design a learnable NI extractor that captures representative features corresponding to normal regions within the test image. Then, we design the NIG decoder to reconstruct local features based on the extracted representative normal information. Finally, defects are segmented by computing the residuals between the reconstructed features and the original defect regions.

\subsection{Normal-Information Extractor}
SEM image of IC exhibits highly intricate backgrounds and pronounced cross-image variability, rendering reliable alignment challenging and consequently degrading segmentation performance. To overcome this limitation, we propose a learnable normal-information extractor. Rather than comparing features drawn from a support set, we directly extract normal information from the test image itself that is representative of its normal regions.

First, pseudo defects and their corresponding binary masks are synthesized on the training image.
Then, a pretrained backbone extracts multi-layer features
\(\{\mathbf{f}_\ell\}_{\ell=1}^{L}\), with each \(\mathbf{f}_\ell \in R^{N \times C}\).
Here, \(N\) denotes the number of patch tokens and \(C\) denotes the feature dimension.

As shown in Figure 2(b), we aggregate features from multiple layers to obtain $\mathbf{F}$, a multi-scale representation that serves as the keys and values of the normal-information extractor. Next, we apply cross-attention by taking $M$ initialized learnable normal tokens $\mathbf{F}_{\text{normal}}^{\text{init}}$ together with $\mathbf{F}$ as input, and obtain $\mathbf{F}_{\text{normal}} \in R^{M \times C}$, which encodes the information of normal regions.The computation is as follows.

\begin{gather*}
\mathbf{F} = \sum_{\ell=1}^{L}\mathbf{f}^{\,\ell},\\
\mathbf{Q} = \mathbf{F}_{\text{normal}}^{\text{init}}\mathbf{W}_{q},\quad
\mathbf{k} = \mathbf{F}\mathbf{W}_{k},\quad
\mathbf{v} = \mathbf{F}\mathbf{W}_{v},\\
\mathbf{F}'_{\text{normal}} = \operatorname{Attention}(\mathbf{Q},\mathbf{k},\mathbf{v}) + \mathbf{F}_{\text{normal}},\\
\mathbf{F}_{\text{normal}} = \operatorname{MLP}(\mathbf{F}'_{\text{normal}}) + \mathbf{F}'_{\text{normal}}.
\end{gather*}

$W_q$, $W_k$, and $W_v \in R^{C \times C}$ denote the learnable parameters in the cross-attention module, and $\mathrm{MLP}$ denotes a multilayer perceptron.

To ensure that the extracted normal information originates from normal regions, we design the loss by first separating normal and pseudo-defect regions using a downsampled mask. We then compute cosine distances between the normal information and the features from normal and abnormal regions, respectively, with the objective of minimizing the distance to normal-region features while maximizing the distance to abnormal-region features. This encourages a strong association between the extracted normal information and truly normal regions. The loss is defined as follows:

\[
\begin{aligned}
d_i^{+} &= \min_{m\in\{1,\ldots,M\}}
\operatorname{distance\_cos}\!\big(\mathbf{F}(i),\,\mathbf{F}_{\text{normal}}^{\,m}\big),\\
d_j^{-} &= \min_{m\in\{1,\ldots,M\}}
\operatorname{distance\_cos}\!\big(\mathbf{F}(j),\,\mathbf{F}_{\text{normal}}^{\,m}\big),\\
\mathcal{L}_{normal} &= \frac{1}{|\Omega_{\mathrm{n}}|}\sum_{i\in\Omega_{\mathrm{n}}} d_i^{+}
+\frac{1}{|\Omega_{\mathrm{a}}|}\sum_{j\in\Omega_{\mathrm{a}}} (1-d_j^{-}),
\qquad \alpha\ge 0,
\end{aligned}
\]

where $\operatorname{distance\_cos}(\cdot,\cdot)$ denotes the cosine similarity between the patch tokens extracted by the pretrained model and the normal-information tokens $\mathbf{F}_{\text{normal}}$. Here, $d_i^{+}$ denotes the distance to the nearest normal patch token for $i\in\Omega_{\mathrm{n}}$, and $d_j^{-}$ denotes the distance to the nearest abnormal patch token for $j\in\Omega_{\mathrm{a}}$.

\subsection{Normal-Information-Guided Decoder}
It is important to note that the extracted normal informations  $\mathbf{F}_{\text{normal}}$ is not a direct surrogate for any single local patch; it encodes a global feature of normality across regions. Consequently, using it to compute patch-level feature distances for anomaly scoring is unreliable. Instead, we leverage its normal origin to guide the reconstruction of IC features. Defects are then segmented by the residual between the reconstructed features and the original features, which suppresses global bias and accentuates fine-grained departures from normal structure.

As shown in Figure 2(c), $\mathbf{F}_{\text{normal}}$ is injected into the decoder to guide feature reconstruction. Because $\mathbf{F}_{\text{normal}}$ represents only normal-region characteristics, we use it as the keys and values. During training, the decoder focuses on reconstructing features from normal regions, which effectively suppresses reconstruction for anomalous queries.

\begin{gather*}
\mathbf{Q}_{\ell} = \mathbf{f}_{\text{decoder}}^{\,\ell-1}\mathbf{W}^{Q}_{\ell},\quad
\mathbf{K}_{\ell} = \mathbf{F}_{\text{normal}}\mathbf{W}^{K}_{\ell},\quad
\mathbf{V}_{\ell} = \mathbf{F}_{\text{normal}}\mathbf{W}^{V}_{\ell},\\
\mathbf{A}_{\ell} = \operatorname{ReLU}\!\big(\mathbf{Q}_{\ell}\mathbf{K}_{\ell}^{\top}\big),
\mathbf{f}_{\text{decoder}}^{\,\ell-1'} = \mathbf{A}_{\ell}\mathbf{V}_{\ell},\quad,\\
\mathbf{f}_{\text{decoder}}^{\,\ell} = \operatorname{MLP}\!\big(\mathbf{f}_{\text{decoder}}^{\,\ell-1'}\big)
+ \mathbf{f}_{\text{decoder}}^{\,\ell-1'} .
\end{gather*}

Here, $\mathbf{f}_{\text{decoder}}^{\,\ell}\in R^{N\times C}$ denotes the output of the $\ell$-th decoder layer, and $\mathbf{W}^{Q}_{\ell}, \mathbf{W}^{K}_{\ell}, \mathbf{W}^{V}_{\ell}\in R^{C\times C}$ are the learnable parameters of the $\ell$-th decoder layer. 

To make the reconstructed normal features closely match the encoder’s original normal features while emphasizing hard regions during backpropagation and suppressing gradient interference from easy regions, we design the loss of reconstruction to directly modulate feature gradients.

\begin{gather*}
w^{\ell}(h,w) = \left[\frac{M^{\ell}(h,w)}{u\!\left(M^{\ell}\right)}\right]^{\gamma},\\[4pt]
\mathcal{L}_{rc} = \frac{1}{L}\sum_{\ell=1}^{L}
\operatorname{distance\_cos}\!\Big(\operatorname{vec}(\mathbf{f}^{\,\ell}),\,
\operatorname{vec}(\hat{\mathbf{f}}_{D}^{\,\ell})\Big),\\[4pt]
\hat{\mathbf{f}}_{D}^{\,\ell}(h,w) = cg\!\big(\mathbf{f}_{D}^{\,\ell}(h,w)\big)\; w^{\ell}(h,w).
\end{gather*}

where $u(M^{\ell})$ represents the average regional cosine distance within a batch, $\gamma \ge 0$ denotes the temperature hyper-parameter, $cg(\cdot)\, w^{\ell}(h,w)$ denotes a gradient adjustment based on the dynamic weight $w^{\ell}(h,w)$, and $\operatorname{vec}(\cdot)$ denotes the flattening operation. The overall training loss of our INP-Former can be expressed as

\[
\mathcal{L}_{\text{total}} \;=\; \mathcal{L}_{rc}\;+\; \lambda\,\mathcal{L}_{normal}.
\]

\section{Experiment}
\label{sec:experiment}
\textbf{Dataset.} In this study, our dataset comprises 2990 SEM images across three process stages: back end of line (BEOL) with 1290 images, deposition (DEP) with 775 images, and dummy poly remove (DPR) with 925 images.

\textbf{Implementation Details:}
We use ViT-Base/14 with DINOv2-R pre-trained weights~\cite{oquab2023dinov2} as the default encoder.
The number of normal tokens extracted by the normal-information extractor is set to $M=6$.
All images are resized to $448\times448$.
Hyperparameters are fixed to $\gamma=3.0$ and $\lambda=0.2$.
We train for 300 epochs with AdamW and a learning rate of $1\times10^{-3}$.
Experiments are conducted on a workstation with four NVIDIA RTX 4090 GPUs.

\subsection{Main Results}
To assess the performance of our method in IC scenarios, we evaluate it across three process stages and compare against state-of-the-art approaches. Image-level AUROC (i-AUROC) and pixel-level AUROC (p-AUROC) results are reported in Table 1.

\begin{table}[h]
\caption{Image-level (i-AUROC) and pixel-level (p-AUROC) on three IC process stages (\%).}
\centering
\tiny
\setlength{\tabcolsep}{6pt}
\begin{tabular}{l c c c c c c}
\noalign{\hrule height 1.2pt}
Method & \multicolumn{2}{c}{BEOL} & \multicolumn{2}{c}{DEP} & \multicolumn{2}{c}{DPR} \\
      & i-AUROC & p-AUROC & i-AUROC & p-AUROC & i-AUROC & p-AUROC \\
\hline
AA-CLIP\cite{ma2025aa} & 95.69 & 93.12 & 97.85 & 96.04 & 73.08 & 89.42 \\
winCLIP\cite{jeong2023winclip}      & 65.08 & 85.62 & 90.86 & 93.25 & 64.69 & 88.24 \\
MUSC\cite{li2024musc}         & 92.24 & 95.04 & 98.01 & 97.02 & 82.77 & 94.12 \\
AnomalyCLIP\cite{zhou2025anomalyclipobjectagnosticpromptlearning}  & 82.10 & 95.40 & 90.90 & 91.10 & 73.70 & 92.16 \\
MAE-IC\cite{lu2023masked}       & 81.70 & 90.20 & 97.60 & 95.20 & 82.60 & 91.20 \\
\textbf{Ours} & \textbf{97.58} & \textbf{96.37} & \textbf{99.19} & \textbf{98.92} & \textbf{97.62} & \textbf{96.62} \\
\noalign{\hrule height 1.2pt}
\end{tabular}
\end{table}

The comparative results show that across the three process stages, our method achieves superior performance on both the i-AUROC and p-AUROC metrics.

\begin{figure}[ht]
        \centering
	\includegraphics[width=0.42\textwidth]{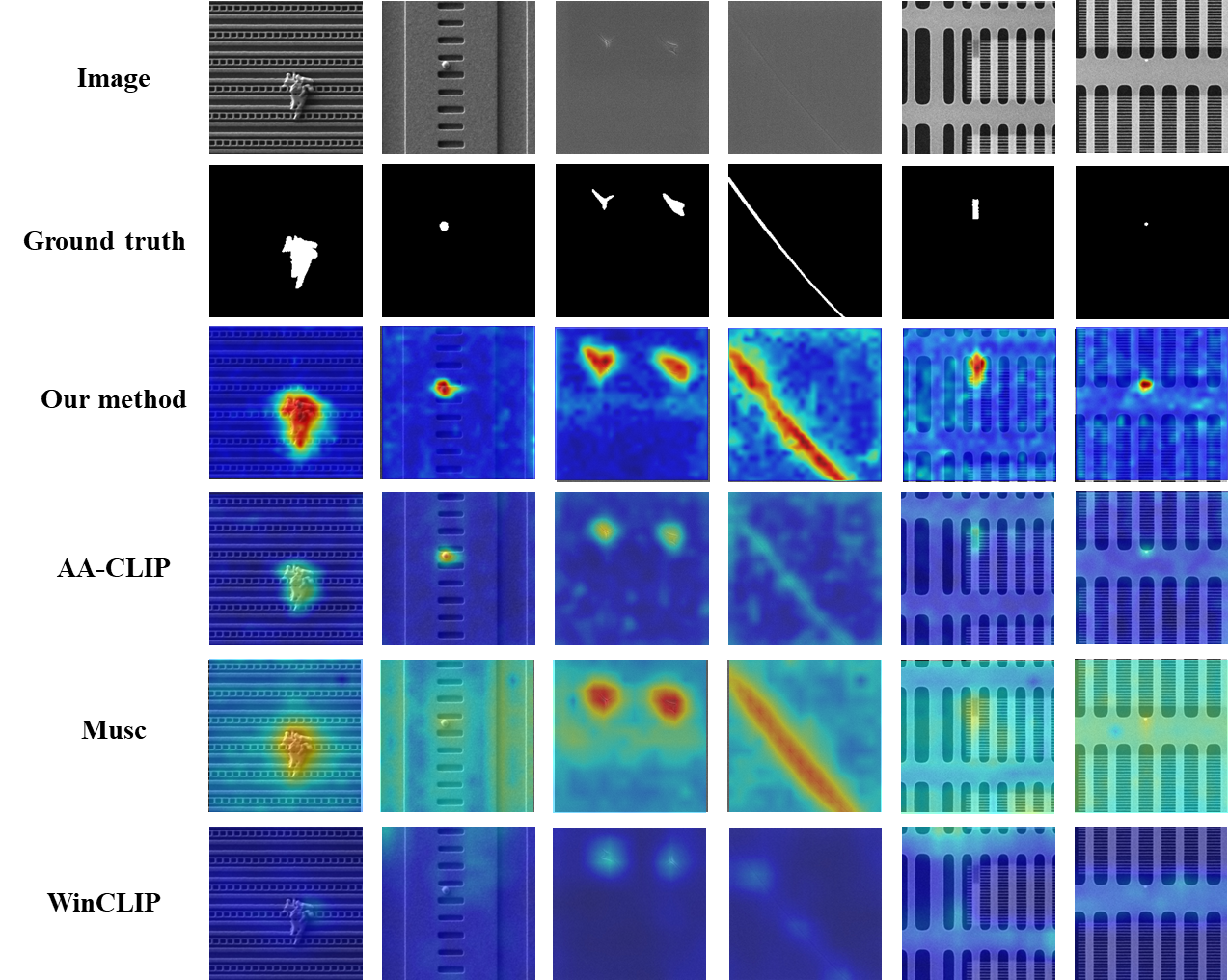}
	\caption{Visualization of different method.}
	\label{Fig:3}       
\end{figure}

In addition, the visual comparison in Fig. 3 shows more precise defect segmentation. We attribute this to two factors: first, despite large-scale pretraining, a notable domain gap remains between generic models and IC SEM imagery; second, many baselines depend on support-set normal features. In contrast, our approach analyzes image-intrinsic normal information within the complex backgrounds of IC SEM images, thereby removing reliance on external support sets and yielding cleaner, more accurate defect masks.

\subsection{Ablation Studies}
We compare defect segmentation using Normal-Information (NI) guided feature reconstruction with a direct matching strategy that compares NI to local features, and we provide visual evidence. As shown in Table 2 and Figure 6, direct matching is more susceptible to noise and NI is not an exact substitute for local normal features. In contrast, the reconstruction-based approach delivers more stable and accurate segmentation in complex backgrounds.

\begin{table}[h]
\caption{Comparison between direct Normal-Information (NI) matching and NI-guided reconstruction on three IC process stages (\%). Metrics are image-level (i-AUROC) and pixel-level (p-AUROC).}
\centering
\tiny
\setlength{\tabcolsep}{6pt}
\begin{tabular}{l c c c c c c}
\noalign{\hrule height 1.2pt}
Method & \multicolumn{2}{c}{BEOL} & \multicolumn{2}{c}{DEP} & \multicolumn{2}{c}{DPR} \\
      & i-AUROC & p-AUROC & i-AUROC & p-AUROC & i-AUROC & p-AUROC \\
\hline
Direct NI matching & 95.82 & 95.41 & 98.79 & 98.12 & 94.67 & 92.51 \\
\textbf{Ours} & \textbf{97.58} & \textbf{96.37} & \textbf{99.19} & \textbf{98.92} & \textbf{97.62} & \textbf{94.62} \\
\noalign{\hrule height 1.2pt}
\end{tabular}
\end{table}

\begin{figure}[ht]
        \centering
	\includegraphics[width=0.42\textwidth]{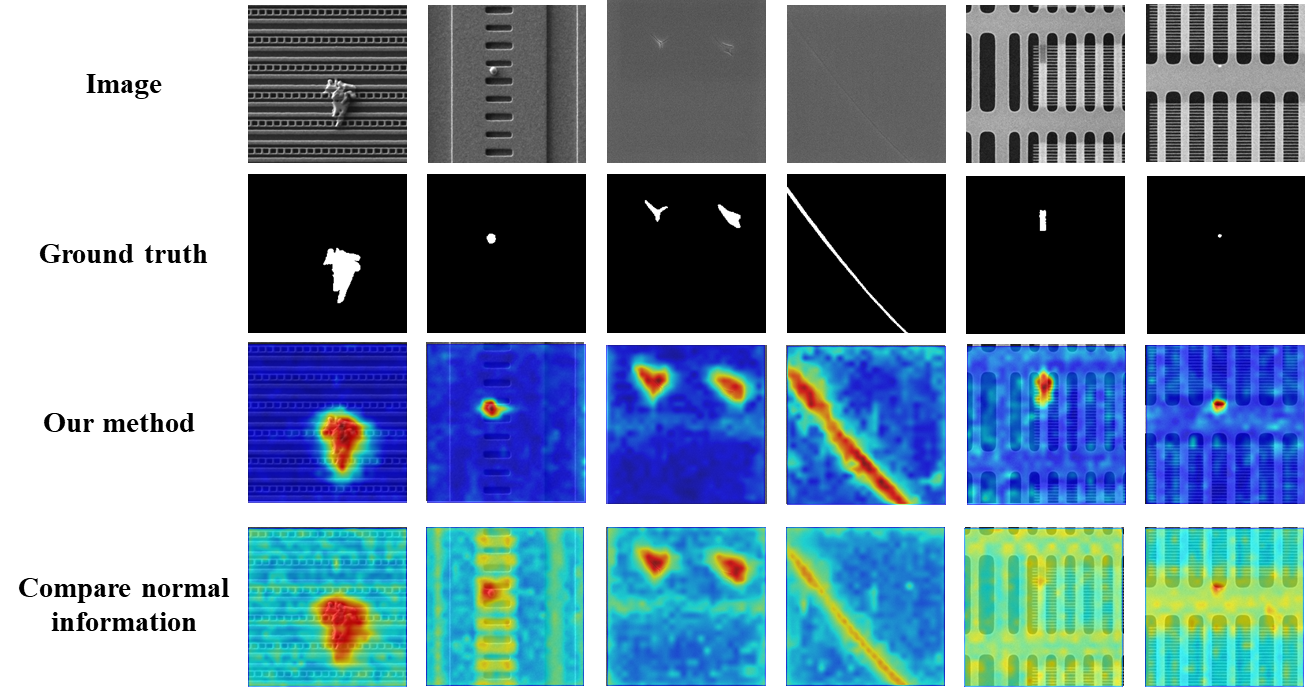}
	\caption{Visual comparison between our normal-information (NI) guided reconstruction method and the baseline that segments defects by directly comparing NI with local features.}
	\label{Fig:4}       
\end{figure}

Model performance under few-shot settings was further evaluated. The results show that our method maintains strong segmentation performance even with limited training samples. 

\begin{table}[h]
\caption{Effect of training sample size (k-shot) on i-AUROC and p-AUROC across three IC process stages.}
\centering
\tiny
\setlength{\tabcolsep}{6pt}
\begin{tabular}{l c c c c c c}
\noalign{\hrule height 1.2pt}
 & \multicolumn{2}{c}{BEOL} & \multicolumn{2}{c}{DEP} & \multicolumn{2}{c}{DPR} \\
 & i-AUROC & p-AUROC & i-AUROC & p-AUROC & i-AUROC & p-AUROC \\
\hline
k=1      & 92.69 & 90.11 & 96.71 & 94.62 & 82.16 & 79.98 \\
k=4      & 95.32 & 92.97 & 97.64 & 96.92 & 92.57 & 89.71 \\
full-shot& 97.58 & 96.37 & 99.19 & 98.92 & 97.62 & 94.62 \\
\noalign{\hrule height 1.2pt}
\end{tabular}
\end{table}

We ablate the number $M$ of normal-information tokens. As shown in Fig. 5, increasing $M$ leads to consistent performance gains that gradually saturate. This suggests that a moderate number of normal tokens helps cover more normal patterns and improves robustness; too few limits representational capacity, whereas too many yields diminishing returns and higher computational cost.

\begin{figure}[ht]
        \centering
	\includegraphics[width=0.4\textwidth]{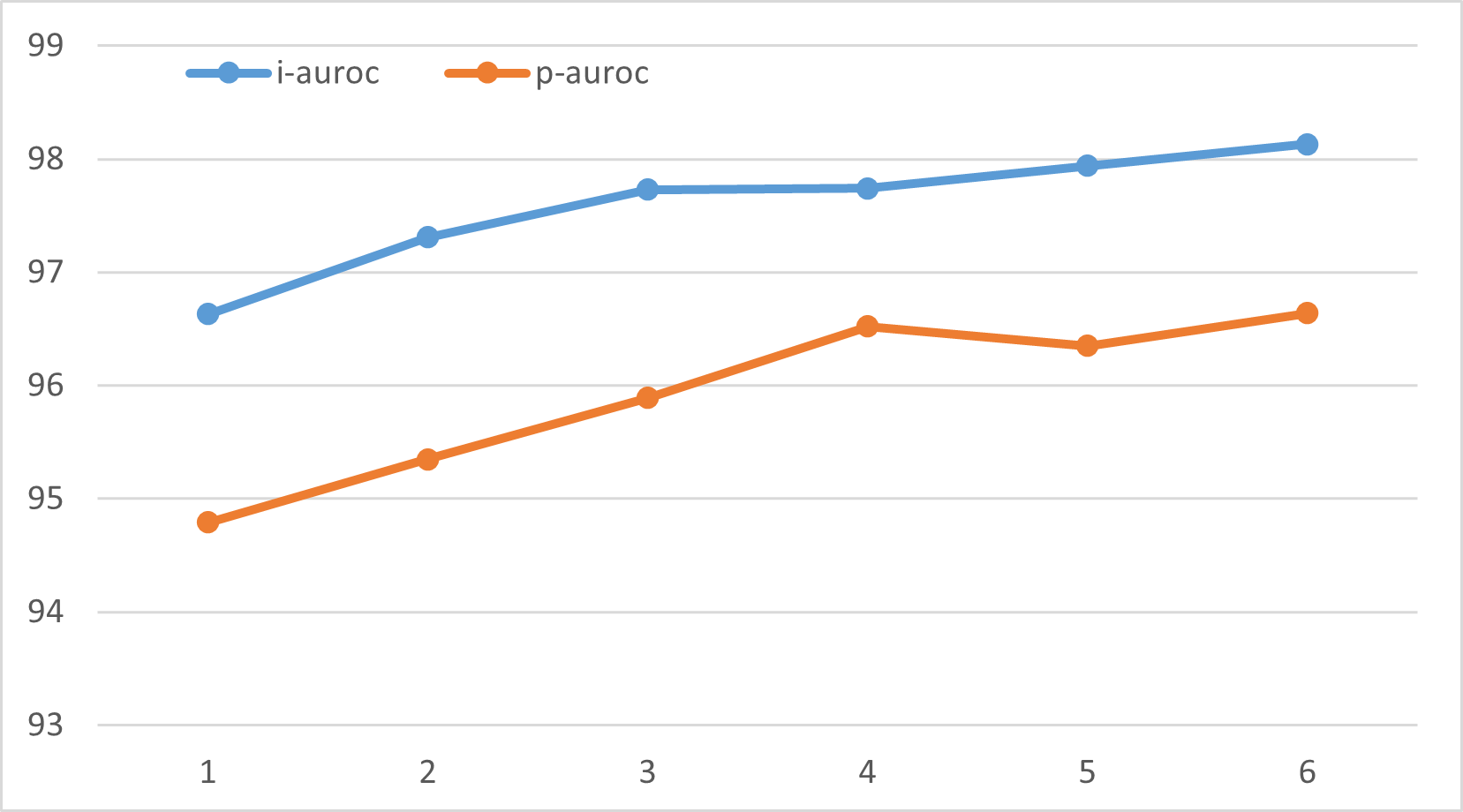}
	\caption{Effect of NI Counts.}
	\label{Fig:5}       
\end{figure}

\section{conclusion}
In this paper, we propose an unsupervised IC defect segmentation framework that does not rely on external normal references by extracting image intrinsic normal information from the test image. A learnable normal information extractor distills stable and representative features of normal regions. Guided by these features, a decoder reconstructs local representations, and the residual between the reconstruction and the original features is used to segment defects under complex SEM backgrounds. Experiments across three process stages show consistent improvements over existing methods on both image level and pixel level metrics, demonstrating robustness to fine grained defects. These results indicate that image intrinsic normality is an effective prior for IC inspection.

\bibliographystyle{IEEEbib}
\bibliography{strings,refs}

\end{document}